\documentclass{midl} % Include author names

\usepackage{mwe} % to get dummy images
\usepackage{float}
\usepackage{floatflt}
\usepackage{caption}
\usepackage{array}
\usepackage{multirow}

\jmlrproceedings{MIDL}{}
\jmlrpages{}
\title[]{Boosting 3D Neuron Segmentation with 2D Vision Transformer Pre-trained on Natural Images} 

\midlauthor{\Name{Yik San Cheng\nametag{$^{1}$}} \Email{yche6976@uni.sydney.edu.au}\\
\Name{Runkai Zhao\nametag{$^{1}$}} \Email{rzha9419@uni.sydney.edu.au}\\
\Name{Heng Wang\nametag{$^{1,2}$}} \Email{heng.wang.cv@gmail.com}\\
\Name{Hanchuan Peng\nametag{$^{3}$}} \Email{h@braintell.org}\\ 
\Name{Weidong Cai\nametag{$^{1}$}} \Email{tom.cai@sydney.edu.au}\\
\addr $^{1}$ School of Computer Science, The University of Sydney, Sydney, Australia\\
\addr $^{2}$ Data61, CSIRO, Sydney, Australia\\
\addr $^{3}$ SEU-ALLEN Joint Center, Institute for Brain and Intelligence, Southeast University, Nanjing, China
  }

\begin{document}

\maketitle

\begin{abstract}
Neuron reconstruction, one of the fundamental tasks in neuroscience, rebuilds neuronal morphology from 3D light microscope imaging data. It plays a critical role in analyzing the structure-function relationship of neurons in the nervous system. However, due to the scarcity of neuron datasets and high-quality SWC annotations, it is still challenging to develop robust segmentation methods for single neuron reconstruction. To address this limitation, we aim to distill the consensus knowledge from massive natural image data to aid the segmentation model in learning the complex neuron structures. Specifically, in this work, we propose a novel training paradigm that leverages a 2D Vision Transformer model pre-trained on large-scale natural images to initialize our Transformer-based 3D neuron segmentation model with a tailored 2D-to-3D weight transferring strategy. Our method builds a knowledge sharing connection between the abundant natural and the scarce neuron image domains to improve the 3D neuron segmentation ability in a data-efficiency manner. Evaluated on a popular benchmark, BigNeuron, our method enhances neuron segmentation performance by 8.71\% over the model trained from scratch with the same amount of training samples. 
\end{abstract}


\section{Introduction}
Single neuron reconstruction aims to extract and digitalize tree-shaped neuron structures from 3D light microscopic images. It is essential to analyze and understand the connectivity and functionality of different neurons within the nervous system ~\cite{zhang2018automated,liu2024single,gao2023single,manubens2023bigneuron,peng2021morphological,qiu2024whole}. Traditional neuron reconstruction methods depend on manual labeling and hand-crafted tracing algorithms, which are labor-intensive and time-consuming ~\cite{wang2021ai}. Recently, learning-based techniques have been developed to extract neuron foreground pixels in a data-driven manner, including multi-scale kernel fusion ~\cite{Wang_2019_CVPR_Workshops}, atrous spatial pyramid pooling ~\cite{li20193d}, global graph reasoning ~\cite{wang2021single}, cross-volume representation learning ~\cite{wang2021voxel},  homogenous model knowledge transfer ~\cite{wang2019segmenting} and 3D point geometry learning ~\cite{zhao2023pointneuron}. Although these works prove that learning-based segmentation models can adaptively learn the complex neuron morphology in an end-to-end fashion, the limited availability of volumetric neuron image datasets with high-quality SWC annotations restricts the segmentation performance. To alleviate this problem, we propose to fully leverage the vast repository of natural image data. By distilling the consensus knowledge from these data, we strive to boost the segmentation capability in accurately interpreting neuronal morphology. Previous studies ~\cite{jiang2018retinal,raj2021automated,mcbee2023pre} in the medical domain have demonstrated that transferring knowledge from 2D natural image data improves segmentation performance. However, utilizing 2D natural knowledge to enhance 3D neuron segmentation models remains unexplored due to the dimension and domain disparity. In this study, we design a novel training paradigm to apply a tailored 2D-to-3D weight transferring strategy on the initialization of Transformer-based 3D neuron segmentation models through the derived knowledge from 2D natural images.  

% However, due to the dimensional disparity between the 2D and 3D domains, the utilization of 2D natural knowledge to enhance 3D neuron segmentation models remains unexplored. 
\begin{figure}[htbp]
 % Caption and label go in the first argument and the figure contents
 % go in the second argument
\floatconts
  {fig:example}
  {\captionsetup{font=footnotesize}\caption{The overview of our proposed training paradigm for 3D neuron segmentation. The network follows an encoder-decoder structure. The 3D neuron image is first divided into several 3D blocks which are then fed to a 3D Vision Transformer (ViT) for slice segmentation. During training phase, the pre-trained weights from a 2D ViT are used to initialize the 3D ViT through a weight transferring strategy. In the end, the segmented slices are stacked together to form the final segmentation prediction which is then forwarded to produce the target SWC file through a neuron tracing method.}}
  {\includegraphics[width=0.95\linewidth]{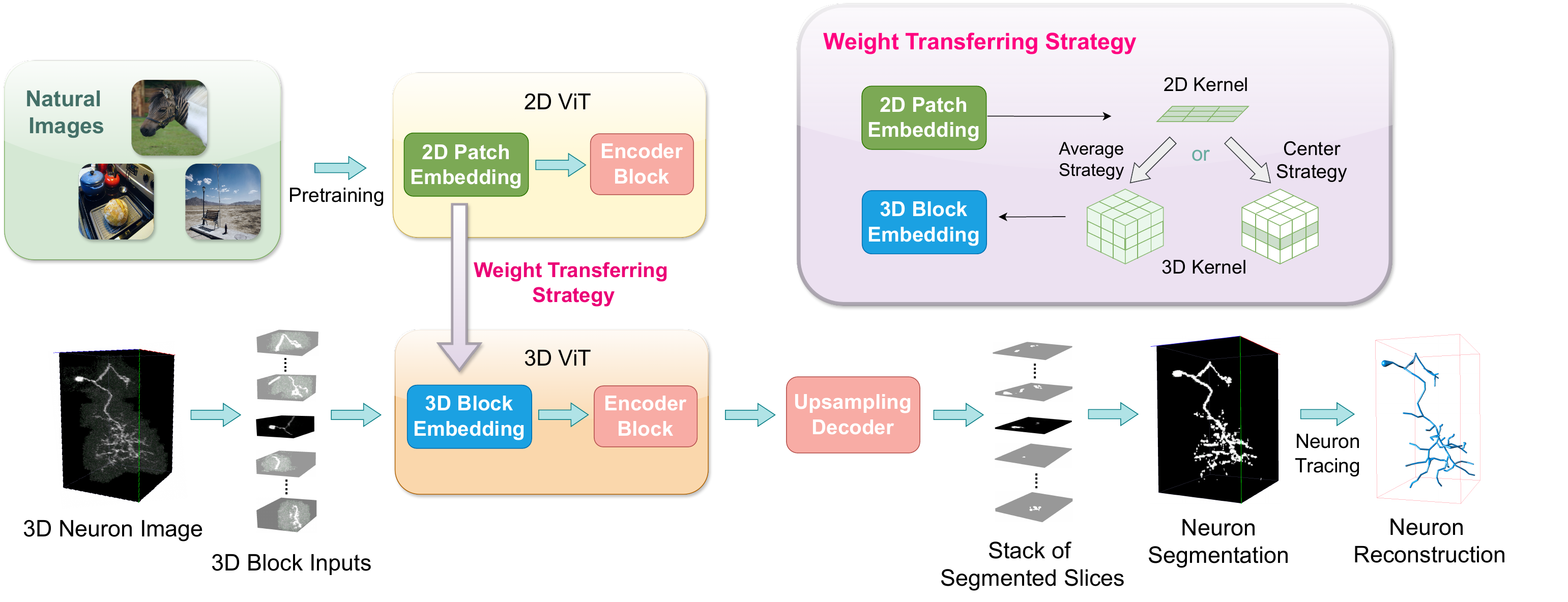}}
  % \label{fig:example} 
\end{figure}

\section{Methodologies}

\subsection{Dataset and Preprocessing}
The neuron datatset utilized in our study is obtained from a publicly accessible Janelia dataset developed by the BigNeuron project ~\cite{peng2015bigneuron,manubens2023bigneuron}. In our study, this dataset is split into training, validation, and testing sets, accounting for 38, 3, and 4 of images respectively. Due to the memory limitation, we divide neuron image volumes into multiple 3D blocks with the size of $100\times100\times5$ and train only with blocks containing salient neuronal voxels (over a predefined foreground ratio). Following the current method ~\cite{Wang_2019_CVPR_Workshops}, we apply the scale-space distance transformation technique to produce neuron segmentation ground truth.  

\subsection{Network Architecture and Weight Transferring Strategy}
Our network architecture is shown in \figureref{fig:example}. We leverage the consensus knowledge from DINO ~\cite{caron2021emerging}, which is a self-supervised 2D Vision Transformer (ViT) model trained on a range of large 2D natural image datasets including ImageNet and COCO. To effectively transfer the prior knowledge grabbed from 2D natural image data to our 3D neuron segmentation model, we propose to apply a weight transferring strategy~\cite{zhang2022adapting} to facilitate the training of our 3D neuron ViT. Specifically, we investigate two variants as shown in \figureref{fig:example}. For the average strategy, we treat all the slices in the block input equally. We duplicate the pre-trained weights of 2D kernels in the 2D patch embedding layer by the number of block depth to create the 3D kernels which are then transferred to the 3D block embedding layer after being divided by the number of the duplication. For the center strategy, we only transfer the pre-trained weights in the 2D kernels to the center slice of the 3D kernels but make neighboring slices to zero. 

% \renewcommand{\arraystretch}{1.3}
% \begin{table}[htbp]
% \footnotesize
% \captionsetup{font=small}
% \caption{Neuron Reconstruction Performance.}
% \centering
% \begin{tabular}{cccccc}
% % \hline
% \hline
% Model & Pre-trained Weights & Transferring Strategy & Input Depth & Mean Dice\uparrow & Mean Hd95\downarrow \\
% \hline
% \hline
% \multicolumn{1}{c}{2D ViT} & \multicolumn{1}{c}{/} & \multicolumn{1}{c}{/} & \multicolumn{1}{c}{1} & \multicolumn{1}{c}{0.4469} & \multicolumn{1}{c}{7.649} \\
% \multicolumn{1}{c}{2D ViT$^{\star}$} & \multicolumn{1}{c}{\checkmark} & \multicolumn{1}{c}{/} & \multicolumn{1}{c}{1} & \multicolumn{1}{c}{\textbf{0.4876}} & \multicolumn{1}{c}{\textbf{2.893}} \\
% \hline
% \multicolumn{1}{c}{3D ViT} & \multicolumn{1}{c}{/} & \multicolumn{1}{c}{/} & \multicolumn{1}{c}{5} & \multicolumn{1}{c}{0.4174} & \multicolumn{1}{c}{37.943} \\
% \multicolumn{1}{c}{3D ViT$^{\star}$} & \multicolumn{1}{c}{\checkmark} & \multicolumn{1}{c}{Average} & \multicolumn{1}{c}{5} & \multicolumn{1}{c}{0.4942} & \multicolumn{1}{c}{2.863} \\
% \multicolumn{1}{c}{3D ViT$^{\star\!\star}$} &	\multicolumn{1}{c}{\checkmark} & \multicolumn{1}{c}{Center} & \multicolumn{1}{c}{5} & \multicolumn{1}{c}{\textbf{0.5045}} & \multicolumn{1}{c}{\textbf{2.810}} \\
% \hline
% \end{tabular}
% \end{table}

% Table 3
\renewcommand{\arraystretch}{1.3}
\begin{table}[htbp]
% \floatconts
%   {tab:example}%
%   {\caption{An Example Table}}%
\footnotesize
\captionsetup{font=small}
\caption{Neuron Segmentation Performance.}
\centering
\begin{tabular}{cccc|cc}
\hline
Model & Pre-trained Weights & Transferring Strategy & Input Depth & Mean Dice$\uparrow$ & Mean Hd95$\downarrow$ \\
\hline
\hline
\multicolumn{1}{c}{\multirow{2}*{2D ViT}} & \multicolumn{1}{c}{/} & \multicolumn{1}{c}{/} & \multicolumn{1}{c|}{1} & \multicolumn{1}{c}{0.4469} & \multicolumn{1}{c}{7.649} \\
~ & \multicolumn{1}{c}{\checkmark} & \multicolumn{1}{c}{/} & \multicolumn{1}{c|}{1} & \multicolumn{1}{c}{\textbf{0.4876}} & \multicolumn{1}{c}{\textbf{2.893}} \\
\hline
\multicolumn{1}{c}{\multirow{3}*{3D ViT}} & \multicolumn{1}{c}{/} & \multicolumn{1}{c}{/} & \multicolumn{1}{c|}{5} & \multicolumn{1}{c}{0.4174} & \multicolumn{1}{c}{37.943} \\
~ & \multicolumn{1}{c}{\checkmark} & \multicolumn{1}{c}{Average} & \multicolumn{1}{c|}{5} & \multicolumn{1}{c}{0.4942} & \multicolumn{1}{c}{2.863} \\
~ & \multicolumn{1}{c}{\checkmark} & \multicolumn{1}{c}{Center} & \multicolumn{1}{c|}{5} & \multicolumn{1}{c}{\textbf{0.5045}} & \multicolumn{1}{c}{\textbf{2.810}} \\
\hline
\label{tab:example}
\end{tabular}
\end{table}

\section{Results and Conclusion}
As presented in Table \ref{tab:example}, leveraging the knowledge extracted from 2D natural images greatly improves the neuron segmentation performance compared with models trained with random initialization, from 0.4174 to 0.5045 in mean Dice and 37.943 to 2.810 in mean 95\% Hausdorff distance (Hd95). It is also found that the center weight transferring strategy achieves the best performance. In addition, including depth information can outperform the slice-to-slice based 2D ViT method. 

In conclusion, our study demonstrates the effectiveness of transferring 2D prior knowledge from natural image data to improve 3D neuron segmentation without introducing additional overhead, which mitigates the data scarcity problem in neuron study. With the same amount of training samples, our 2D-to-3D weight transferring training paradigm can boost the neuron segmentation performance by 8.71\% on the BigNeuron benchmark.

\bibliography{midl-samplebibliography}

\begin{thebibliography}{19}
\providecommand{\natexlab}[1]{#1}
\providecommand{\url}[1]{\texttt{#1}}
\expandafter\ifx\csname urlstyle\endcsname\relax
  \providecommand{\doi}[1]{doi: #1}\else
  \providecommand{\doi}{doi: \begingroup \urlstyle{rm}\Url}\fi

\bibitem[Caron et~al.(2021)Caron, Touvron, Misra, J{\'e}gou, Mairal, Bojanowski, and Joulin]{caron2021emerging}
Mathilde Caron, Hugo Touvron, Ishan Misra, Herv{\'e} J{\'e}gou, Julien Mairal, Piotr Bojanowski, and Armand Joulin.
\newblock Emerging properties in self-supervised vision transformers.
\newblock In \emph{Proceedings of the IEEE/CVF international conference on computer vision}, pages 9650--9660, 2021.

\bibitem[Gao et~al.(2023)Gao, Liu, Wang, Wu, Gou, and Yan]{gao2023single}
Le~Gao, Sang Liu, Yanzhi Wang, Qiwen Wu, Lingfeng Gou, and Jun Yan.
\newblock Single-neuron analysis of dendrites and axons reveals the network organization in mouse prefrontal cortex.
\newblock \emph{Nature Neuroscience}, 26\penalty0 (6):\penalty0 1111--1126, 2023.

\bibitem[Jiang et~al.(2018)Jiang, Zhang, Wang, and Ko]{jiang2018retinal}
Zhexin Jiang, Hao Zhang, Yi~Wang, and Seok-Bum Ko.
\newblock Retinal blood vessel segmentation using fully convolutional network with transfer learning.
\newblock \emph{Computerized Medical Imaging and Graphics}, 68:\penalty0 1--15, 2018.

\bibitem[Li and Shen(2019)]{li20193d}
Qiufu Li and Linlin Shen.
\newblock 3d neuron reconstruction in tangled neuronal image with deep networks.
\newblock \emph{IEEE transactions on medical imaging}, 39\penalty0 (2):\penalty0 425--435, 2019.

\bibitem[Liu et~al.(2024)Liu, Gao, Chen, and Yan]{liu2024single}
Sang Liu, Le~Gao, Jiu Chen, and Jun Yan.
\newblock Single-neuron analysis of axon arbors reveals distinct presynaptic organizations between feedforward and feedback projections.
\newblock \emph{Cell Reports}, 43\penalty0 (1), 2024.

\bibitem[Manubens-Gil et~al.(2023)Manubens-Gil, Zhou, Chen, Ramanathan, Liu, Liu, Bria, Gillette, Ruan, Yang, et~al.]{manubens2023bigneuron}
Linus Manubens-Gil, Zhi Zhou, Hanbo Chen, Arvind Ramanathan, Xiaoxiao Liu, Yufeng Liu, Alessandro Bria, Todd Gillette, Zongcai Ruan, Jian Yang, et~al.
\newblock Bigneuron: a resource to benchmark and predict performance of algorithms for automated tracing of neurons in light microscopy datasets.
\newblock \emph{Nature Methods}, 20\penalty0 (6):\penalty0 824--835, 2023.

\bibitem[McBee et~al.(2023)McBee, Moradinasab, Brown, and Syed]{mcbee2023pre}
Payden McBee, Nazanin Moradinasab, Donald~E Brown, and Sana Syed.
\newblock Pre-training segmentation models for histopathology.
\newblock In \emph{Medical Imaging with Deep Learning, short paper track}, 2023.

\bibitem[Peng et~al.(2015)Peng, Hawrylycz, Roskams, Hill, Spruston, Meijering, and Ascoli]{peng2015bigneuron}
Hanchuan Peng, Michael Hawrylycz, Jane Roskams, Sean Hill, Nelson Spruston, Erik Meijering, and Giorgio~A Ascoli.
\newblock Bigneuron: large-scale 3d neuron reconstruction from optical microscopy images.
\newblock \emph{Neuron}, 87\penalty0 (2):\penalty0 252--256, 2015.

\bibitem[Peng et~al.(2021)Peng, Xie, Liu, Kuang, Wang, Qu, Gong, Jiang, Li, Ruan, et~al.]{peng2021morphological}
Hanchuan Peng, Peng Xie, Lijuan Liu, Xiuli Kuang, Yimin Wang, Lei Qu, Hui Gong, Shengdian Jiang, Anan Li, Zongcai Ruan, et~al.
\newblock Morphological diversity of single neurons in molecularly defined cell types.
\newblock \emph{Nature}, 598\penalty0 (7879):\penalty0 174--181, 2021.

\bibitem[Qiu et~al.(2024)Qiu, Hu, Huang, Gao, Wang, Wang, Ren, Shi, Chen, Wang, et~al.]{qiu2024whole}
Shou Qiu, Yachuang Hu, Yiming Huang, Taosha Gao, Xiaofei Wang, Danying Wang, Biyu Ren, Xiaoxue Shi, Yu~Chen, Xinran Wang, et~al.
\newblock Whole-brain spatial organization of hippocampal single-neuron projectomes.
\newblock \emph{Science}, 383\penalty0 (6682):\penalty0 eadj9198, 2024.

\bibitem[Raj et~al.(2021)Raj, Londhe, and Sonawane]{raj2021automated}
Ritesh Raj, Narendra~D Londhe, and Rajendra Sonawane.
\newblock Automated psoriasis lesion segmentation from unconstrained environment using residual u-net with transfer learning.
\newblock \emph{Computer Methods and Programs in Biomedicine}, 206:\penalty0 106123, 2021.

\bibitem[Wang et~al.(2019{\natexlab{a}})Wang, Zhang, Song, Liu, Huang, Chen, Peng, and Cai]{Wang_2019_CVPR_Workshops}
Heng Wang, Donghao Zhang, Yang Song, Siqi Liu, Heng Huang, Mei Chen, Hanchuan Peng, and Weidong Cai.
\newblock Multiscale kernels for enhanced u-shaped network to improve 3d neuron tracing.
\newblock In \emph{Proceedings of the IEEE/CVF Conference on Computer Vision and Pattern Recognition (CVPR) Workshops}, June 2019{\natexlab{a}}.

\bibitem[Wang et~al.(2019{\natexlab{b}})Wang, Zhang, Song, Liu, Wang, Feng, Peng, and Cai]{wang2019segmenting}
Heng Wang, Donghao Zhang, Yang Song, Siqi Liu, Yue Wang, Dagan Feng, Hanchuan Peng, and Weidong Cai.
\newblock Segmenting neuronal structure in 3d optical microscope images via knowledge distillation with teacher-student network.
\newblock In \emph{2019 IEEE 16th International Symposium on Biomedical Imaging (ISBI 2019)}, pages 228--231. IEEE, 2019{\natexlab{b}}.

\bibitem[Wang et~al.(2021{\natexlab{a}})Wang, Song, Tang, Zhang, Yu, Liu, Zhang, Liu, and Cai]{wang2021ai}
Heng Wang, Yang Song, Zihao Tang, Chaoyi Zhang, Jianhui Yu, Dongnan Liu, Donghao Zhang, Siqi Liu, and Weidong Cai.
\newblock Ai-enhanced 3d biomedical data analytics for neuronal structure reconstruction.
\newblock In \emph{Humanity Driven AI: Productivity, Well-being, Sustainability and Partnership}, pages 135--163. Springer, 2021{\natexlab{a}}.

\bibitem[Wang et~al.(2021{\natexlab{b}})Wang, Song, Zhang, Yu, Liu, Pengy, and Cai]{wang2021single}
Heng Wang, Yang Song, Chaoyi Zhang, Jianhui Yu, Siqi Liu, Hanchuan Pengy, and Weidong Cai.
\newblock Single neuron segmentation using graph-based global reasoning with auxiliary skeleton loss from 3d optical microscope images.
\newblock In \emph{2021 IEEE 18th International Symposium on Biomedical Imaging (ISBI)}, pages 934--938. IEEE, 2021{\natexlab{b}}.

\bibitem[Wang et~al.(2021{\natexlab{c}})Wang, Zhang, Yu, Song, Liu, Chrzanowski, and Cai]{wang2021voxel}
Heng Wang, Chaoyi Zhang, Jianhui Yu, Yang Song, Siqi Liu, Wojciech Chrzanowski, and Weidong Cai.
\newblock Voxel-wise cross-volume representation learning for 3d neuron reconstruction.
\newblock In \emph{Machine Learning in Medical Imaging: 12th International Workshop, MLMI 2021, Held in Conjunction with MICCAI 2021, Strasbourg, France, September 27, 2021, Proceedings 12}, pages 248--257. Springer, 2021{\natexlab{c}}.

\bibitem[Zhang et~al.(2018)Zhang, Liu, Song, Feng, Peng, and Cai]{zhang2018automated}
Donghao Zhang, Siqi Liu, Yang Song, Dagan Feng, Hanchuan Peng, and Weidong Cai.
\newblock Automated 3d soma segmentation with morphological surface evolution for neuron reconstruction.
\newblock \emph{Neuroinformatics}, 16:\penalty0 153--166, 2018.

\bibitem[Zhang et~al.(2022)Zhang, Huang, Zhou, Lungren, and Yeung]{zhang2022adapting}
Yuhui Zhang, Shih-Cheng Huang, Zhengping Zhou, Matthew~P Lungren, and Serena Yeung.
\newblock Adapting pre-trained vision transformers from 2d to 3d through weight inflation improves medical image segmentation.
\newblock In \emph{Machine Learning for Health}, pages 391--404. PMLR, 2022.

\bibitem[Zhao et~al.(2023)Zhao, Wang, Zhang, and Cai]{zhao2023pointneuron}
Runkai Zhao, Heng Wang, Chaoyi Zhang, and Weidong Cai.
\newblock Pointneuron: 3d neuron reconstruction via geometry and topology learning of point clouds.
\newblock In \emph{Proceedings of the IEEE/CVF Winter Conference on Applications of Computer Vision}, pages 5787--5797, 2023.

\end{thebibliography}

\end{document}